\documentclass[journal]{IEEEtran}
\pdfoutput=1
\usepackage[turkish]{babel}
\usepackage[utf8]{inputenc} 
\usepackage[T1]{fontenc}

\usepackage{cite}

\ifCLASSINFOpdf
  \usepackage[pdftex]{graphicx}
\else
\fi
\usepackage{subcaption}

\usepackage{cite}
\usepackage{amsfonts}
\usepackage{algorithm}
\usepackage{booktabs}
\usepackage{array}
\usepackage{multirow}
\usepackage{graphicx}
\usepackage{listings}
\usepackage[utf8]{inputenc}
\usepackage[T1]{fontenc}
\usepackage{url}
\usepackage{hyperref}

\begin{document}


%
\title{Derin Pekiştirmeli Öğrenmede Özellik Çıkarma için Atari Oyunlarında Üretici Çekişmeli Ağların Kullanımı\\
Using Generative Adversarial Nets on Atari Games for Feature Extraction in Deep Reinforcement Learning\\
}

\author{Ayberk~Aydın and Elif~Surer\\
Department of Modeling and Simulation, Graduate School of Informatics, Middle East Technical University, Ankara, Turkey\\
{\tt\small aydin.ayberk@metu.edu.tr, elifs@metu.edu.tr}

}


\maketitle
%

\begin{abstract}
Deep Reinforcement Learning (DRL) has been successfully applied in several research domains such as robot navigation and automated video game playing. However, these methods require excessive computation and interaction with the environment, so enhancements on sample efficiency are required. The main reason for this requirement is that sparse and delayed rewards do not provide an effective supervision for representation learning of deep neural networks. In this study, Proximal Policy Optimization (PPO) algorithm is augmented with Generative Adversarial Networks (GANs) to increase the sample efficiency by enforcing the network to learn efficient representations without depending on sparse and delayed rewards as supervision. The results show that an increased performance can be obtained by jointly training a DRL agent with a GAN discriminator.
\end{abstract}


\begin{abstract}
Derin Pekiştirmeli Öğrenme, robot navigasyonu ve otomatikleştirilmiş video oyunu oynama gibi araştırma alanlarında başarıyla uygulanmaktadır. Ancak, kullanılan yöntemler ortam ile fazla miktarda etkileşim ve hesaplama gerektirmekte ve bu nedenle de örnek verimliliği yönünden iyileştirmelere ihtiyaç duyulmaktadır. Bu gereksinimin en önemli nedeni, gecikmeli ve seyrek ödül sinyallerinin derin yapay sinir ağlarının etkili betimlemeler öğrenebilmesi için yeterli bir denetim sağlayamamasıdır. Bu çalışmada, Proksimal Politika Optimizasyonu algoritması Üretici Çekişmeli Ağlar (ÜÇA) ile desteklenerek derin yapay sinir ağlarının seyrek ve gecikmeli ödül sinyallerine bağımlı olmaksızın etkili betimlemeler öğrenmesi teşvik edilmektedir. Elde edilen sonuçlar önerilen algoritmanın örnek verimliliğinde artış elde ettiğini göstermektedir.
\end{abstract}


%

\section{G{\footnotesize İ}r{\footnotesize İ}ş}

Derin öğrenme yöntemlerinin görüntü sınıflandırma, görüntü segmentasyonu ve görsel nesne tespiti gibi alanlardaki başarısının altında derin yapay sinir ağlarının öznitelik çıkarma becerileri yatmaktadır. Derin pekiştirmeli öğrenme (Deep Reinforcement Learning - DRL) ise derin yapay sinir ağları kullanılarak bir pekiştirmeli öğrenme ajanının yüksek boyutlu ham görsel veri üzerinde başarıyla eğitilmesinden sonra \cite{Mnih2015} önemli bir araştırma alanı haline gelmiştir. 

DRL, yakın zamandaki gelişmeler ile Atari, Go, Satranç oyunları ve robot kolu kontrolü gibi görevlerde insanüstü performans gösterme seviyesine ulaşmıştır. Ancak, bu yöntemler hala oldukça fazla örneğe ihtiyaç duymaktadır ve bu nedenle de verimsiz olarak değerlendirilmektedirler. Örneğin, DRL metotları, Atari oyunlarında insan seviyesine ulaşmak için onlarca milyon adım boyunca eğitime ihtiyaç duymaktadır.

Denetimli algoritmalar için denetimsiz verilerden yararlanmak, öğrenme algoritmalarının başarısını yükseltmektedir ve veri etiketleme, veri toplamaya kıyasla oldukça zorlu bir iştir. Bu yüzden denetimsiz veriden yararlanmak, algoritma performansını iyileştirmek için etkili bir yöntemdir. Aynı durum pekiştirmeli öğrenme için de geçerlidir. Bu alanda öğrenen ajanın elde ettiği ödül, denetim sinyali olarak kullanılmakla birlikte, seyrek ve gecikmeli olarak elde edilmektedir. Verimli bir derin öğrenme sistemi, eğer denetim sinyali sık, gecikmesiz ve güvenilir değil ise öznitelik ve betimlemeleri denetim sinyaline ihtiyaç duymadan çıkarabilmelidir.

Bu çalışmada, DRL algoritmaları denetimsiz öğrenme yöntemleri ile desteklenerek kullanılan ağların denetimsiz bir şekilde etkili betimlemeler (representation) öğrenmesi sağlanmış ve örnek verimliliğinin artırılması üzerinde çalışılmıştır. Bu amaçla, denetimsiz öğrenme literatüründe sıkça kullanılan Üretici Çekişmeli Ağlar \cite{Goodfellow} ve otokodlayıcıla r\cite{Hinton} ile deneyler yapılmıştır ve iki yöntemin verdiği sonuçlar iki farklı oyun üzerinde (Pong ve Breakout) incelenmiştir. 
\section{İlg{\footnotesize İ}l{\footnotesize İ} Çalışmalar}

Derin öğrenmede girdi betimlemelerini denetim olmaksızın öğrenmek aktif bir araştırma konusudur ve denetimsiz/özdenetimli öğrenme yöntemlerinde iki ana paradigma bulunmaktadır. Bunlardan birincisi Otokodlayıcı (Autoencoder) sınıfı altındadır. Otokodlayıcılar, girdi olarak aldığı veriyi önce daha düşük bir boyuta sıkıştırıp ardından bu sıkıştırılmış betimleme ile orijinal girdiyi tekrar oluşturmaya çalışır. Böylelikle otokodlayıcılar girdi verisini düşük bir boyutta ifade etmeyi girdideki örüntüleri kodlayarak öğrenebilir. Öğrenme tamamlandıktan sonra sıkıştırılmış verinin, girdinin düşük boyutlu bir betimlemesi olması beklenmektedir. Denetimsiz betimleme öğrenmede diğer bir paradigma ise Üretici Çekişmeli Ağlar (ÜÇA; Generative Adversarial Networks - GANs) sınıfındadır. ÜÇA'ların ana fikri, biri Üretici (Generator) ve biri Ayrıştırıcı (Discriminator) olmak üzere iki adet yapay sinir ağının tümleşik bir biçimde eğitilmesi ve eğitimin sonunda veri kümesine benzeyen gerçekçi örneklerin elde edilmesine dayanır. Eğitim sırasında üretici, gerçekçi örnekler üretmeye çalışırken ayrıştırıcı bu örneklerin gerçek olup olmadığını anlamaya çalışır. Değişimli bir şekilde sırayla eğitilen bu ağlardan ayrıştırıcı, eğitimin sonunda verinin düşük boyutlu betimlemelerini öğrenebilmektedir \cite{Radford}. ÜÇA'ları evrişimli derin yapay sinir ağları ile eğitebilen ilk çalışma olan DCGAN (Deep Convolutional GANs) \cite{Radford}, aynı zamanda CIFAR-10 veri kümesinin betimlemelerini denetimsiz bir şekilde öğrenmeyi başarmıştır. Bu betimlemeleri lineer bir sınıflandırıcı ile çok az sayıda etiketlenmiş veri ile sınıflandırıp mevcut en iyi performansı elde etmiştir.

Yüksek boyutlu ham veriler ile başarılı bir şekilde pekiştirmeli öğrenme ajanı eğitmek, 2015'te Mnih ve arkadaşlarının \cite{Mnih2015} Deep Q-Networks'u (DQN) literatüre kazandırdıkları çalışmaya kadar mümkün olmamıştır. Bu çalışmada yazarlar deneyim tekrarı (experience replay) ve hedef ağ (target network) kullanarak Atari oyunlarının yalnızca pikseller ile insanüstü bir seviyede oyun oynamayı öğrenmesini sağlamışlardır. Double DQN \cite{VanHasselt}, Asynchronous Advantage Actor Critic (A3C) \cite{Mnih2016} gibi algoritmalar ile DRL algoritmalarının kararlılık ve örnek verimliliği artırılmıştır. Birbirinden bağımsız bu geliştirmelerin bir kısmı birlikte kullanılarak çok daha yüksek bir örnek verimliliği elde edilebilmektedir \cite{Hessel}. 
Ayrıca, Schulman ve arkadaşları \cite{Schulman}, Proksimal Politika Eniyilemesi (Proximal Policy Optimization - PPO) ile doğrudan politika eniyileme için oldukça basit bir objektif fonksiyonu tanımlayarak Atari ve robot kontrolü simülasyonu alanında o tarihe kadar olan en yüksek performansı elde etmiştir.

Bahsedilen yöntemler DRL'nin kararlılığını ve örnek verimliliğini artırsa da, ajanların gözlemlerinin verimli betimlemelerini öğrenme konusuna odaklanmamıştır. Betimleme öğrenme için denetimsiz öğrenmenin pekiştirmeli öğrenme yöntemleri ile birlikte uygulanabileceğini araştıran ilk çalışmada \cite{Lange}, otokodlayıcı ön eğitimi ile öznitelik çıkarımı sağlanmıştır ancak iyileştirmeler küçük ölçekli problemlerden öteye geçememiştir. Kimura \cite{Kimura} bir otokodlayıcıyı rastgele politika ile elde edilmiş örnekler ile ön eğitimden geçirerek gerçek kameralar ile çekilen Taş-Kağıt-Makas oyununda öğrenilen betimlemeleri kullanarak pekiştirmeli eğitim yapmıştır ancak performansta kararlı bir iyileşme gözlemlenememiştir. Özetle, literatürdeki çalışmalar otokodlayıcıların betimleme öğrenmede yeterli iyileştirmeler sağlayamadıklarını göstermektedir. 

Bu çalışmada ise Atari oyunları üzerinde otokodlayıcılar ile yapılan deneyler  görselleştirilmiş, otokodlayıcıların neden bu alanda yeterli iyileştirmeler sağlayamadıkları sorusuna açıklama getirilmiş ve ÜÇA çerçevesinin bu problemi çözmekte başarılı olduğu sonuçlar ile desteklenerek gösterilmiştir.

\section{Yöntem}
\subsection{Eğitim Ortamı}
Bu çalışmada eğitim ortamı olarak Arcade Learning Environment \cite{Bellemare} kullanılmıştır. Bu ortam, Atari oyunları üzerinde Pekiştirmeli Öğrenme algoritmalarının denenmesi için oluşturulmuş olup, birçok Atari oyunu içermektedir. Bu çalışmada Pong ve Breakout video oyunları üzerinde deneyler yapılmıştır.

\subsection{Veri Önişleme}
Önişleme için her üç karede bir aksiyon seçilerek üç kare boyunca tekrarlanmıştır ve elde edilen üç kare gri skalaya çevrilip art arda konulmuştur. Bu veri, ardından 64x64 boyutuna indirilerek [0-255] aralığından [0-1] aralığına çekilmiş ve ağa girdi olarak verilmiştir. Genellikle Atari üzerinde yapılan deneylerde 84x84 büyüklüğünde bir girdi kullanılsa da bu çalışmada üç kanallı girdi alan DCGAN mimarisine sadık kalmak amacıyla üç kanallı 64x64 boyutunda bir gözlem kullanılmıştır. ÜÇA için kayıp fonksiyonu olarak Göreceli Ortalama En Küçük Kare (Relativistic Average Least Squares) kaybı\cite{Jolicoeur} kullanılmıştır.

\subsection{Pekiştirmeli Öğrenme ve Betimleme Öğrenme}
Pekiştirmeli Öğrenme alanında başarıyla uygulanan birçok yöntem bulunsa da, bu çalışmada, kolayca uygulanabilmesi ve yüksek performansı nedeniyle PPO \cite{Schulman} algoritması ve betimleme öğrenme ile özellik çıkarma (feature extraction) yöntemi olarak ÜÇA \cite{Goodfellow} kullanılmıştır. Bu algoritmalar için kullanılan üst değişkenler (hyperparameter) \cite{Schulman} Tablo Tablo 1'de gösterilmektedir. 

\subsection{Yapay Sinir Ağı Mimarisi}
Pekiştirmeli öğrenme mimarisi olarak paylaşımlı parametrelere sahip bir Aktör-Kritik mimarisi kullanılmıştır. Bu mimaride dört adet sıralı evrişimsel katman ve son katmana bağlı biri Aktör, biri Kritik olmak üzere iki adet tamamen bağlı (fully connected) çıktı katmanı bulunmaktadır. Aktör katmanı, mümkün olan her aksiyon için bir olasılık vermekte olup, Kritik katmanı, içinde bulunulan durumun değer fonksiyonunu tahmin etmektedir. Bu mimarideki her iki çıktının değeri ve kaybı, ortamdaki etkileşimlerden elde edilen ödüllere bağlıdır. Bu ödüller çoğu zaman gecikmeli ve seyrek olarak elde edilmektedir ve bu durum kullanılan yapay sinir ağı için verimli betimlemelerin öğrenilmesini zorlaştırmaktadır. Yapay sinir ağının etkili betimlemeler öğrenebilmesi amacıyla, bu Aktör-Kritik mimarisi son evrişimsel katmana eklenen bir adet tamamen bağlı Ayrıştırıcı katmanı ile genişletilmiştir. Bu sayede ayrıştırıcının öğrendiği betimlemeler Aktör-Kritik ağı tarafından kullanılabilmektedir. Bu katmanın amacı, gerçek örnekleri sahte örneklerden ayırt etmektir. Bu nedenle, ağa hem Üretici tarafından üretilen sahte gözlemler hem de bulunulan ortamdan elde edilmiş gerçek gözlemler verilmektedir. Ayrıştırıcı (Discriminator) ile genişletilmiş bu yeni Aktör-Kritik (Actor-Critic) mimarisine "Actor-Critic-Discriminator" (ACD) adı verilmiştir. Belirtilmelidir ki, bu çalışma, Kostrikov'un "Discriminator Actor-Critic" adlı çalışmasıyla \cite{Kostrikov} isim benzerliği taşımaktadır ancak uygulanan yöntem ve çözülen problem olarak iki çalışma birbirinden oldukça farklıdır. Kostrikov'un çalışması, Taklit Öğrenme (Imitation Learning) üzerine bir çalışma olup kullanılan ayrıştırıcı, insan tarafından oluşturulmuş politikalar ile yapay politikalar arasında ayrım yapmayı amaçlamaktadır ve ÜÇA ayrıştırıcısı ile bir ilgisi bulunmamaktadır. Kostrikov'un çalışmasındaki ayrıştırıcının girdisi insan tarafından üretilen girdiler veya yapay politikalar iken, bu çalışmadaki ayrıştırıcının girdisi üretici ağın çıktıları ve ajan gözlemleridir. Ayrıca, Kostrikov'un çalışmasındaki ayrıştırıcı, bu çalışmanın aksine paylaşılan parametreler ile betimleme veya öznitelik öğrenmek için kullanılmamaktadır.

\begin{table}[t!]
    \centering
     \caption{\textsc{Kullanılan üst değ{\footnotesize İ}şkenler{\footnotesize İ}n değerler{\footnotesize İ}}}
     \begin{tabular}{|c|c|c|}
     \hline
     Değişken & Değer & Açıklama \\
     \hline
     $\gamma$ & 0.99 & Geciken ödül için azalma \\
     $\lambda$ & 0.95 & Genellenmiş avantaj tahmini\\
     $\epsilon$ & 0.1 & PPO kayıp kırpma oranı\\
     Yığın Büyüklüğü & 32 & Tek seferde ağa giren veri sayısı \\
     Eniyileyici & RMSProp & Eniyileme algoritması \\
     Öğrenme Hızı & 0.0003 & Eniyileme güncelleme faktörü \\
     Paralel Ortam Sayısı & 8 & Paralel ortam sayısı\\
     T & 128 & Ortamın elde ettiği gözlem sayısı\\
     Dönem (Epoch) Sayısı & 3 & Verilerin üzerinden geçme sayısı\\
     Gizli Uzay Boyutu & 100 & Üretici ağın girdi boyutu\\
     C1 & 1.0 & PPO ilke kaybı katsayısı\\
     C2 & 0.01 & PPO entropi kaybı katsayısı\\
     \hline
     \end{tabular}
    \label{table:hyperparameters}
\end{table}

Üretici Çekişmeli Ağlar'ı evrişimli derin sinir ağları ile kullanabilmek için Radford ve arkadaşlarının \cite{Radford} çalışmasındaki prensipler takip edilmiştir. Bu doğrultuda, ACD ağında Leaky Relu aktivasyonu, Üretici ağında Relu aktivasyonu ve Yığın Normalleştirme (Batch Normalization) kullanılmıştır ve yaygın olarak kullanılan örnekleme yöntemi max pooling yerine evrişimli katmanlarda ikili kaydırma kullanılmıştır. Evrişimli ağın her katmanında aynı boyutta kernel (4x4), kaydırma (2) ve dolgulama (1) kullanılmış olup, ilk katmanda üç kanalı 64 kanala yükseltip diğer katmanlarda kanal sayısını girdisinin iki katına çıkaran dört katman kullanılmıştır ve son evrişimli katmana üç adet paralel çıktı katmanı konulmuştur. Üretici mimarisi ise transpoz evrişimsel katmanlardan \cite{Dumoulin} oluşmakta ve parametreleri bu mimarinin simetriği olmakla birlikte, Relu aktivasyonu ve Yığın Normalleştirme bulundurmaktadır.

\section{Deneysel Sonuçlar}
\subsection{Üretilen Otokodlayıcı ve ÜÇA örnekleri}
\begin{figure}[t!]
  \shorthandoff{=} 
  \begin{subfigure}{0.48878\linewidth}
  \includegraphics[width=\linewidth, height=1.5in]{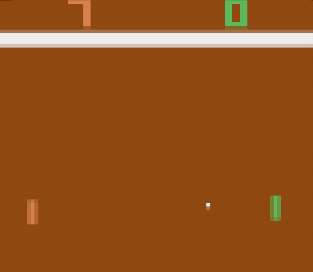}
  \end{subfigure} 
  \begin{subfigure}{0.495\linewidth}
  \includegraphics[width=\linewidth, height=1.5in]{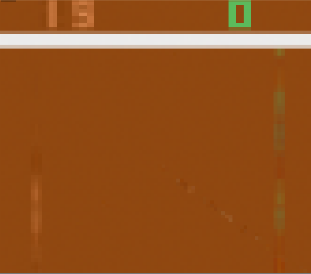}
  \end{subfigure}
  \begin{subfigure}{0.495\linewidth}
  \includegraphics[width=\linewidth, height=1.5in, trim={0.25cm 0.25cm 0.25cm 0.25cm}]{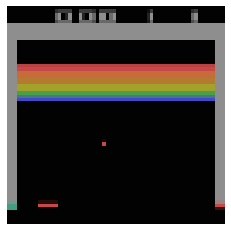}
  \end{subfigure} 
  \begin{subfigure}{0.495\linewidth}
  \includegraphics[width=\linewidth, height=1.5in, trim={0.25cm 0.25cm 0.25cm 0.25cm}]{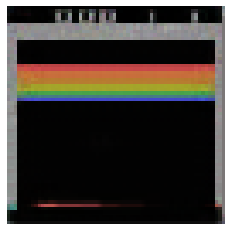}
  \end{subfigure}
  
  \shorthandon{=} 
  \caption{Pong ve Breakout oyunu ekran çerçeveleri (solda) ve otokodlayıcı geri dönüşümleri (sağda). Bu geri dönüşümlerde top ve raket pozisyonu gibi önemli özniteliklerin geri dönüştürülemediği gözlemlenmektedir.}
  \label{fig:ae_pong_breakout}
\end{figure}

\begin{figure}[t!]
  \shorthandoff{=} 
  \begin{subfigure}{0.49\linewidth}
  \includegraphics[width=\linewidth, height=1.5in]{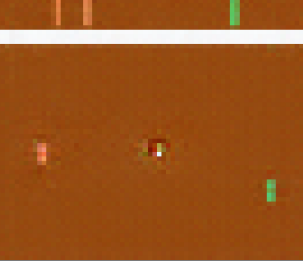}
  \end{subfigure}
  \begin{subfigure}{0.49\linewidth}
  \includegraphics[width=\linewidth, height=1.5in]{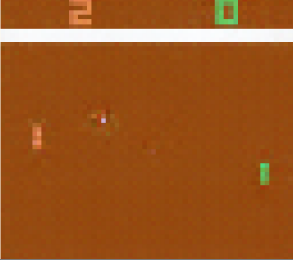}
  \end{subfigure}
  
  \shorthandon{=} 
  \caption{Pong oyunu çerçevelerinde eğitilmiş ÜÇA Üretici çıktıları. Otokodlayıcının aksine, bahsedilen öznitelikler keskin bir biçimde çıkarılabilmektedir.}
  \label{fig:gan_pong}
\end{figure}

\begin{figure}[t!]
  \shorthandoff{=} 
  \begin{subfigure}{0.49\linewidth}
  \includegraphics[width=\linewidth, height=1.5in]{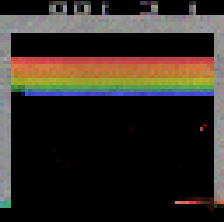}
  \end{subfigure}
  \begin{subfigure}{0.49\linewidth}
  \includegraphics[width=\linewidth, height=1.5in]{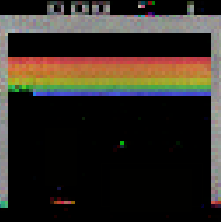}
  \end{subfigure}
  
  \shorthandon{=} 
  \caption{Breakout oyunu çerçevelerinde eğitilmiş ÜÇA Üretici çıktıları. Otokodlayıcının aksine, bahsedilen öznitelikler keskin bir biçimde çıkarılabilmektedir.}
  \label{fig:gan_breakout}
\end{figure}

Bu bölümde Pong oyunu çerçeveleri (frame) üzerinde evrişimsel otokodlayıcı ve DCGAN ile elde edilmiş sonuçlar ve örnekler gösterilmektedir. Şekil \ref{fig:ae_pong_breakout} üzerinde otokodlayıcı ile elde edilen yeniden oluşturulan örnek çerçeveler gösterilmektedir. Şekil \ref{fig:gan_pong} ve \ref{fig:gan_breakout} üzerinde ise ÜÇA ile oluşturulmuş örnekler yer almaktadır.

\subsection{ACD ile Derin Pekiştirmeli Öğrenme}
\begin{figure}[t!]
  \shorthandoff{=} 
  \begin{subfigure}{\linewidth}
  \includegraphics[width=\linewidth, trim={0.1cm 0.1cm 0.1cm 0.1cm}, clip]{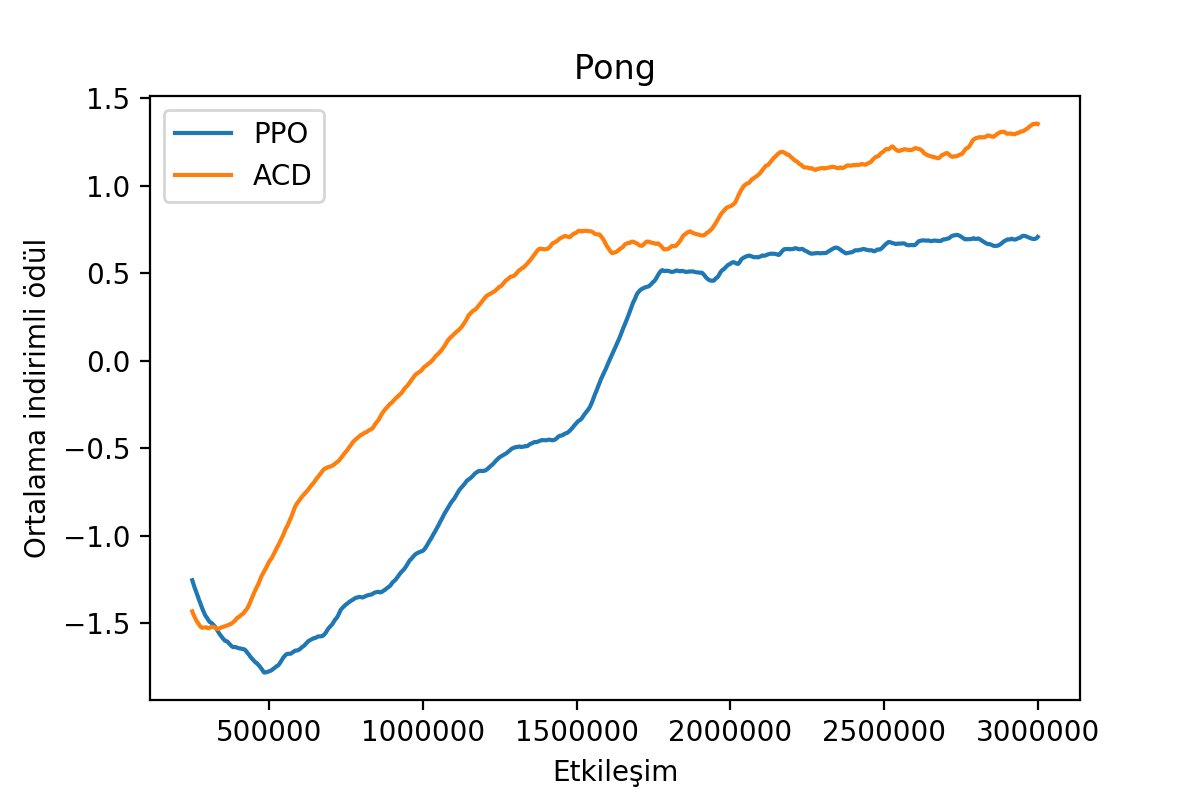}
  \end{subfigure}
  \shorthandon{=} 
  \caption{Pong oyunu üzerinde eğitim sırasında etkileşim başına elde edilen ortalama indirimli ödüller.}
  \label{fig:plot_pong}
\end{figure}

\begin{figure}[t!]
  \shorthandoff{=} 
  \begin{subfigure}{\linewidth}
  \includegraphics[width=\linewidth, trim={0.1cm 0.1cm 0.1cm 0.1cm}, clip]{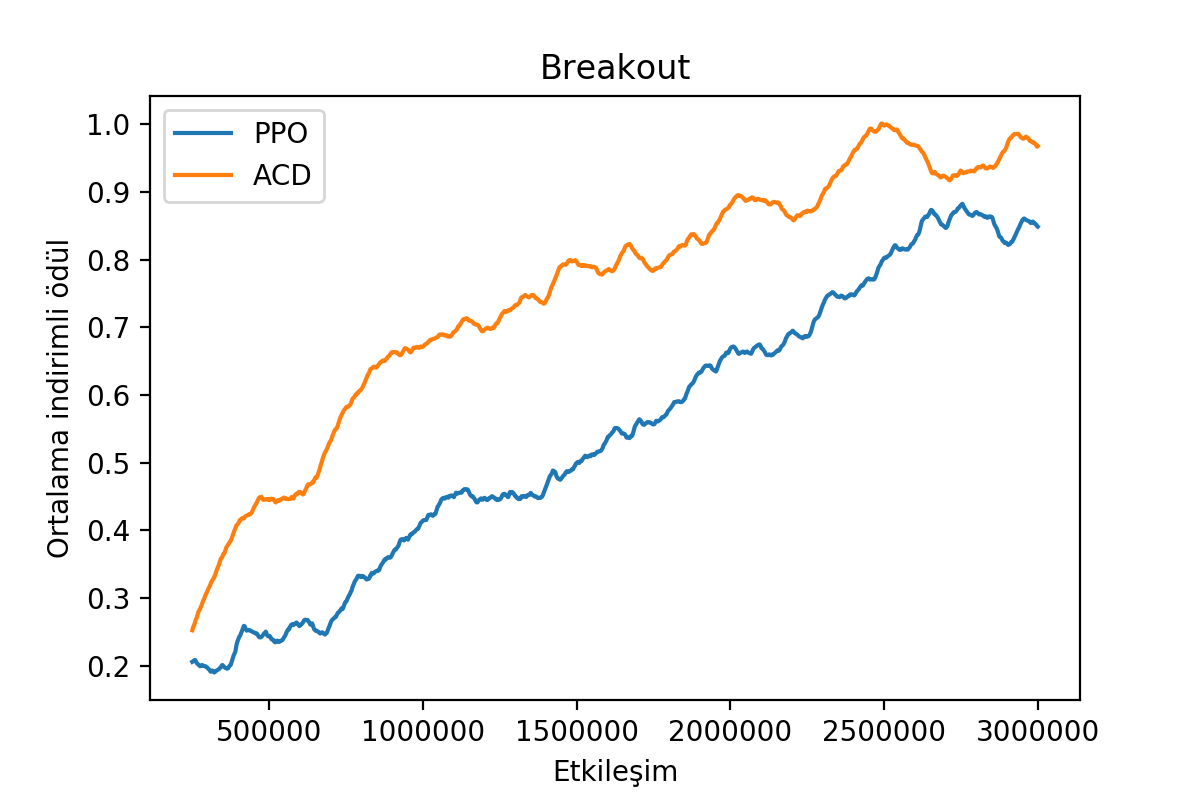}
  \end{subfigure}
  \shorthandon{=} 
  \caption{Breakout oyunu üzerinde eğitim sırasında etkileşim başına elde edilen ortalama indirimli ödüller.}
  \label{fig:plot_breakout}
\end{figure}

Şekil \ref{fig:plot_pong} ve \ref{fig:plot_breakout} üzerinde PPO algoritması ile ve bu algoritmaya ÜÇA kayıp fonksiyonu eklenmiş versiyonu (ACD); Pong ve Breakout oyunlarında üç milyon çerçevelik etkileşim boyunca eğitilirken eğitim sırasında son 100 karede elde edilen ortalama indirimli ödüller gösterilmiştir. Tutarlılık amacıyla, her algoritma üç kere eğitilip sonuçların ortalaması alınmıştır.

\section{Tartışma}
Şekil \ref{fig:ae_pong_breakout} üzerinde görüldüğü gibi otokodlayıcı, ortalama çerçeve piksel uzaklığını eniyilemek adına arkaplan rengi, sabit şekiller gibi özellikleri tamamen geri döndürüp top ve raket gibi hareketli objeleri bulanık bir şekilde geri döndürmektedir. Bunun nedeni, otokodlayıcılarda kullanılan ortalama kare piksel uzaklığı kayıp fonksiyonudur. Bu kayıp fonksiyonu kullanıldığında iki görüntü arasındaki uzaklığın, aynı konumdaki piksellerin değerleri arasındaki uzaklığın ortalama değeri olduğu yönünde bir model varsayımında bulunulmasıdır. 

Otokodlayıcılar, düşük boyutlu koddan orijinal girdiyi geri çevirirken bu uzaklığı küçülttüğü için eniyilenmiş modellerde dahi ortaya bulanık geri dönüşümler çıkmaktadır. ÜÇA ise bu kayıp fonksiyonu yerine öğrenilen çekişmeli kayıp fonksiyonu kullanarak bahsedilen sorunu çözebilmektedir. 

Şekil \ref{fig:gan_pong} ve \ref{fig:gan_breakout} üzerinde ÜÇA ile elde edilmiş örnekler gösterilmektedir. Bu şekillerde görülmektedir ki, pekiştirmeli öğrenme ağının seçeceği aksiyon için kullanması gereken önemli öznitelikler otokodlayıcı tarafından kaybedilirken ÜÇA tarafından keskin bir şekilde çıkarılabilmektedir. Bu sonuçlar, daha önceki çalışmalarda otokodlayıcı ile öğrenilen betimlemelerin Atari üzerindeki pekiştirmeli öğrenme görevlerinde neden etkili olamadığını açıklamaktadır. Şekil \ref{fig:plot_pong} ve \ref{fig:plot_breakout} üzerinde görüldüğü üzere, Pong ve Breakout oyunlarında üç milyon etkileşim süresince eğitilen ÜÇA kaybı ile desteklenen ACD mimarisi, PPO algoritmasından daha iyi bir performans göstermektedir. 

Elde edilen sonuçlar gelişime açık olmakla birlikte, farklı mimariler, A3C veya deneyim tekrarı bulunduran DQN türevleri gibi farklı pekiştirmeli öğrenme algoritmaları ile geliştirilebilir. 

\section{Sonuç}
Bu çalışmada DRL yöntemlerinin örnek verimliliğinin ÜÇA ile desteklenerek artırılması incelenmiştir ve bunun için diğer çalışmalarda denenmiş otokodlayıcı ile desteklemenin neden etkisiz olduğu görsel verilerle desteklenerek açıklanmıştır. Yapılan deneylerde, ÜÇA kaybı ile genişletilmiş PPO algoritmasının, PPO algoritmasından tutarlı bir şekilde daha iyi performans gösterdiği gözlemlenmiştir. Ayrıca, otokodlayıcı ile yapılan geri dönüştürme deneylerinde, eniyilenen otokodlayıcıların piksel bazında ortalama uzaklık ile eğitilmeleri sebebiyle gerekli öznitelikleri öğrenemediği gözlemlenmiştir.

\end{document}